\documentclass[lettersize,journal]{IEEEtran}
\usepackage{amsmath,amsfonts}
\usepackage{algorithmic}
\usepackage{array}
\usepackage[caption=false,font=normalsize,labelfont=sf,textfont=sf]{subfig}
\usepackage{textcomp}

\usepackage{tabularx}
\usepackage{booktabs}
\newcolumntype{Y}{>{\centering\arraybackslash}X}  % 定义 Y 列：可伸缩 + 居中
\usepackage{stfloats}
\usepackage{url}
\usepackage{verbatim}
\usepackage{graphicx}
\usepackage{booktabs}   % \toprule, \midrule, \bottomrule
\usepackage{multirow}   % \multirow
\usepackage[hidelinks]{hyperref}
\def\BibTeX{{\rm B\kern-.05em{\sc i\kern-.025em b}\kern-.08em
    T\kern-.1667em\lower.7ex\hbox{E}\kern-.125emX}}
\usepackage{balance}
\begin{document}
\title{Robotic Scene Cloning:Advancing Zero-Shot Robotic Scene Adaptation in Manipulation via Visual Prompt Editing}

\author{Binyuan Huang$^{1, \dag}$, Yuqing Wen$^{2, \dag}$, Yucheng Zhao$^{3, \ddag}$, Yaosi Hu$^{4}$, Tiancai Wang$^{3}$, \\ Chang Wen Chen$^{4}$, Haoqiang Fan$^{3}$ and Zhenzhong Chen$^{1}$
\thanks{$^{\dag}$: This work was done during the internship at Dexmal.}% <-this % stops a space
\thanks{$^{\ddag}$: Project lead.}%
\thanks{$^{1}$ Binyuan Huang and Zhenzhong Chen are with Wuhan University.}%
\thanks{$^{2}$ Yuqing Wen is with University of Science and Technology of China.}%
\thanks{$^{3}$ Yucheng Zhao, Tiancai Wang and Haoqiang Fan are with Dexmal.}%
\thanks{$^{4}$ Yaosi Hu and Chang Wen Chen are with The Hong Kong Polytechnic University.}%
}

\maketitle

\markboth{Journal of \LaTeX\ Class Files,~Vol.~18, No.~9, September~2020}%
{How to Use the IEEEtran \LaTeX \ Templates}

\begin{abstract}
Modern robots can perform a wide range of simple tasks and adapt to diverse scenarios in the well-trained environment. 
However, deploying pre-trained robot models in real-world user scenarios remains challenging due to their limited zero-shot capabilities, often necessitating extensive on-site data collection. To address this issue, we propose Robotic Scene Cloning (RSC), a novel method designed for scene-specific adaptation by editing existing robot operation trajectories. RSC achieves accurate and scene-consistent sample generation by leveraging a visual prompting mechanism and a carefully tuned condition injection module. 
Not only transferring textures but also performing moderate shape adaptations in response to the visual prompts, RSC demonstrates reliable task performance across a variety of object types.
Experiments across various simulated and real-world environments demonstrate that RSC significantly enhances policy generalization in target environments. 
\end{abstract}


\section{Introduction}

\begin{figure*}[t]
    \centering
              \includegraphics[width=1.0\textwidth]{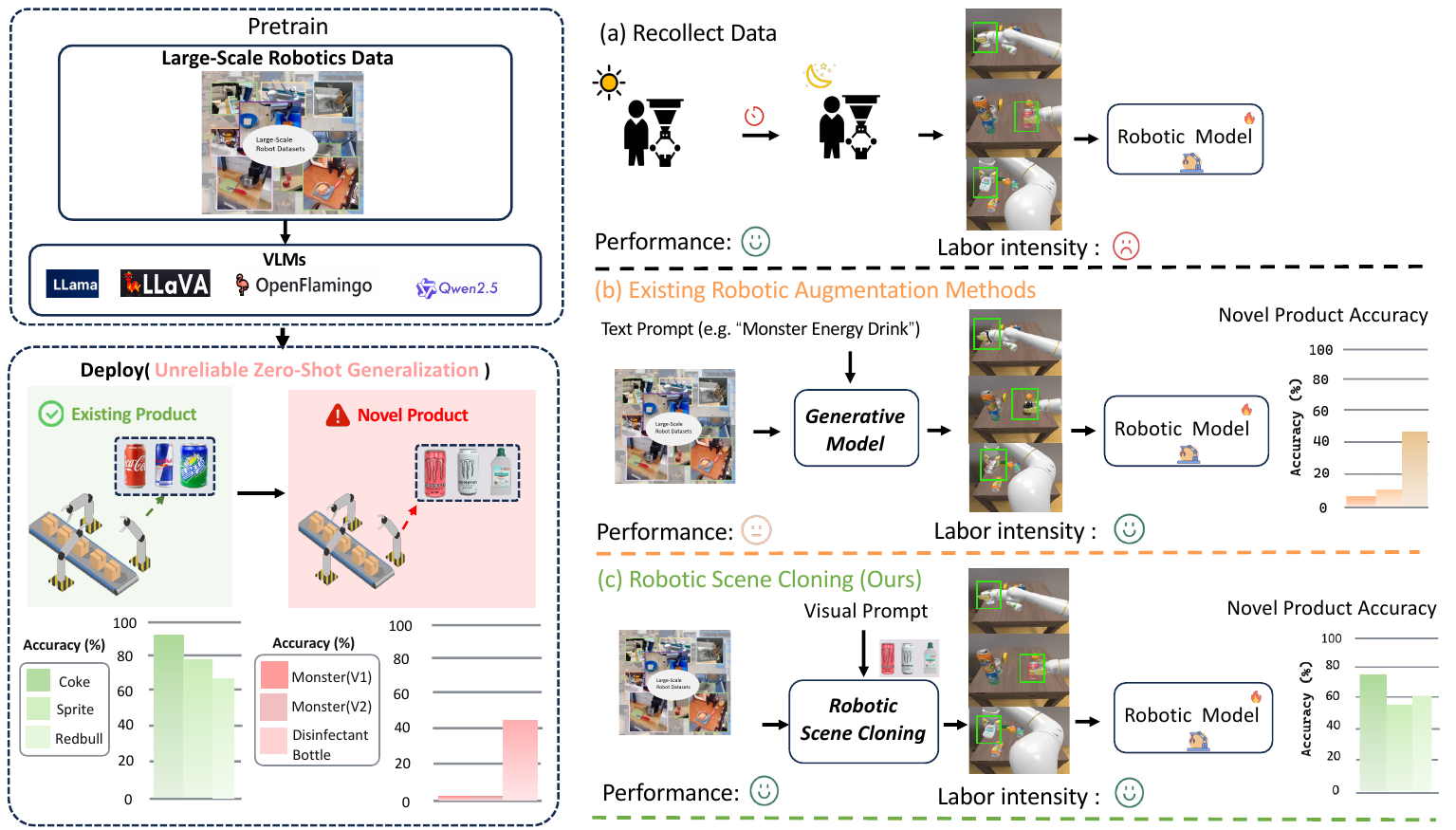}
    \caption
{
 Existing robotic policies face challenges when migrating from training to deployment environments, particularly in handling novel products. (a) Recollect Data: Collecting deployment-specific data enables fine-tuning for high accuracy but is labor-intensive and moderately data-efficient. (b) Existing Embodied Augmentation Methods: Augmentation using text prompts reduces labor but achieves limited accuracy and low data efficiency. (c) Robotic Scene Cloning: Cloning scenes with visual cues from the deployment environment achieves high accuracy with better data efficiency and lower labor cost. The comparison highlights the trade-offs in accuracy, labor intensity, and data efficiency for each method.
The two rightmost bar charts show the success rates of grasping Monster Energy Drink (V1), Monster Energy Drink (V2), and Disinfectant Bottle in the Simpler environment under distractor conditions.
}
\label{fig:intro}
\end{figure*}

In recent years, robotic models have made remarkable advancements across diverse tasks\cite{openvla,rt1,rt2,diffusionpolicy,gr1,r3m,octo,openworld}, However, even the most advanced models often struggle to achieve reliable zero-shot performance when deployed in real-world scenarios. For example, the state-of-the-art vision-language model CogACT\cite{cogact} achieves a success rate of 89.6\% on a pre-trained task (e.g., grasping a Coke bottle) but fails to maintain adequate performance when the Coke is replaced with a Disinfectant Bottle, as illustrated in Fig. \ref{fig:intro}. This simple example simulates a practical situation where new products or objects are introduced into the robot's deployment environment, such as a customer’s home or a industrial product line, highlighting a significant limitation of current robotics models.

A traditional solution to this issue involves collecting new data for the specific environment and fine-tuning the robotic models. However, robotic data collection is labor-intensive and time-consuming. For instance, Google reported that assembling 13K RT-1\cite{rt1} samples required 17 months and 13 robots. An alternative approach is data augmentation, which aims to make better use of existing data. Methods like ROSIE\cite{roise}, GenAug\cite{genaug}, and GreenAug\cite{greenaug} generate diverse samples using text-conditioned diffusion models, improving model performance. However, the visually unbounded nature of these samples provides only limited benefits for specific deployment environments. Similarly, traditional rule-based augmentation techniques, such as color jitter and random cropping, are ineffective for addressing this challenge.
More recently, RoboTransfer~\cite{robotransfer} introduces
video diffusion models for robot data augmentation, leveraging depth,
surface normals, and object-level attributes to synthesize visual textures and reduce the texture gap in both sim-to-real
and real-to-real transfer.

In this paper, we propose a novel robotic data synthesis method called Robotic Scene Cloning (RSC) to effectively address the zero-shot deployment challenge. Unlike other data augmentation approaches, our method specifically targets the deployment environment, prioritizing precise data synthesis over general-purpose diversity. 
% As illustrated in solution (b) of Fig. \ref{fig:intro}, existing augmentation approaches focus on generating diverse samples using text-based conditions; however, these samples often lack consistency with the deployment environment. 
As illustrated in solution (b) of Fig.\ref{fig:intro}, existing augmentation
approaches focus on generating diverse samples using text-based
conditions; however, these samples often misalign with the deployment
environment at the object level, since the generated objects do
not match the target products.
In contrast, as shown in solution (c) of Fig. \ref{fig:intro}, our method replicates the precise visual conditions of the deployment environment. These synthesized samples closely resemble real-world collected data, significantly enhancing model performance in the target environment.

The key feature of Robotic Scene Cloning lies in its ability to generate
accurate and scene-consistent samples that go beyond simple texture
replacement by adapting both appearance and shape to the target
deployment objects. 
To achieve this, we developed a visual instruction-guided image editing framework tailored specifically for robotic scene cloning. This framework incorporates three robotics-specific capabilities: (1) precise placement of visual prompts to ensure accurate object positioning, (2) preservation of semantic consistency in non-edited regions to maintain task-relevant context, and (3) depth-consistent editing to support valid manipulation sequences. These features enable RSC to produce high-fidelity synthetic samples that closely replicate real-world conditions, ensuring reliable policy transfer to target environments.

We evaluate the effectiveness of our method in deployment environments, encompassing both reproducible simulation settings and practical real-world scenarios. To demonstrate its advantages, we compare Robotic Scene Cloning with state-of-the-art generative robotic data augmentation methods. In the SIMPLER environment, RSC outperforms GreenAug\cite{greenaug} by achieving an 35\% higher success rate in the novel object setting and shows a 42.5\% improvement over the zero-shot pre-trained model. In the CALVIN environment\cite{calvin}, RSC achieves an average sequence length of 2.57 in the novel robotic scene setting(with unseen background and foreground), significantly surpassing the baseline model's average sequence length of 1.79. Furthermore, in real-world experiments, RSC demonstrates adaptability to cross-shape cloning, as well as to single-object and multi-object trajectories, achieving a 30\% performance improvement for robotic policies on unseen multi-object long-horizon and single-object short-horizon tasks, which have no real-world demonstrations.
This is enabled by reusing a single original
trajectory to synthesize multiple deployment-specific trajectories
(e.g., converting a grasping trajectory for a banana into ones for
grasping a cube or a glue stick), thereby  improving the
reusability and utilization of collected demonstrations.

In summary, our key contributions are:
\begin{itemize}
    \item We propose a novel data synthesis method called Robotic Scene Cloning, designed to effectively address the zero-shot deployment challenge for robotic models. This approach is distinguished by its visual prompting mechanism, allowing precise control over both the generated objects, including moderate shape variations, and the background environments.
    \item Experimental results demonstrate significant performance improvements in various language-conditioned visuomotor policies when using our method, showcasing its clear advantages over existing generative augmentation approaches.
\end{itemize}

\section{Related Work}
\subsection{Language-conditioned Embodied Policies}
Recent advancements in language-conditioned embodied policies have greatly enhanced robots' ability to execute complex tasks by interpreting natural language commands and visual cues\cite{openvla,cogact,robo1,robo2,robo3,robo4,cliport,peract}. Models like CLIPort\cite{cliport} and PerAct\cite{peract}, leveraging CLIP\cite{clip}, have shown strong performance in robotic manipulation. Recent advanced models like OpenVLA\cite{openvla}, a 7B-parameter open-source vision-language-action model trained on 970k real-world robot demonstrations, and CogACT\cite{cogact}, which employs a componentized architecture with a diffusion-based action module, enhance generalization and fine-tuning efficiency for new tasks. However, both struggle with knowledge transfer to highly novel scenarios, limiting their adaptability.

\subsection{Generative Data Augmentation in Embodied Intelligence.}
Generative models, particularly diffusion models, have advanced robotic learning by enabling synthetic augmentation, reducing reliance on real-world data collection. Early methods like ROSIE\cite{roise} and GenAug\cite{genaug} used text-to-image models for dataset augmentation. ROSIE employed diffusion-based inpainting to integrate unseen objects and backgrounds, improving generalization. GenAug introduced depth-guided visual diversity, enhancing robustness to new environments.
Recent approaches, such as CACTI\cite{mandi2022cacti} and GreenAug\cite{greenaug}, adopted alternative strategies. CACTI combined generative models with expert data for task-specific augmentations like kitchen object manipulation. 
GreenAug employs chroma keying to replace green-screen backgrounds with diverse textures and uses generative inpainting in GreenAug-Gen to add realistic scenes like kitchens or living rooms, enhancing visual generalization with a simpler, effective approach.
RoboEngine~\cite{roboengine} segments the robot and task objects and
uses a fine-tuned diffusion model to regenerate physics- and
task-aware backgrounds, providing plug-and-play background-level
augmentation to improve visual generalization from single-scene
demonstrations.
our visual-prompt-based framework targets
deployment-specific adaptation: using deployment images as visual
prompts, it edits both foreground and background objects in robot
demonstrations, not only transferring textures but also performing
moderate shape adaptations aligned with the visual prompts.

\subsection{Image translation for robotics}
Recent advancements in image translation for robotics have utilized both GAN-based and diffusion-based methods to address the sim-to-real gap. GAN-based approaches like RL-CycleGAN\cite{rl-cyclegan} adapted CycleGAN\cite{cyclegan} with scene consistency losses, ensuring that simulated images retain key semantic and spatial features critical for downstream tasks during domain adaptation. Diffusion-based methods, in contrast, focus on precise control over image generation. ALDM-Grasping\cite{aldm} employed a layout-to-image diffusion process, enabling photorealistic transformations while preserving spatial layout and object relationships critical for robotic manipulation. Similarly, LucidSim\cite{lucidsim} utilized depth-conditioned ControlNet to achieve geometrically consistent scene generation, effectively bridging the sim-to-real gap for diverse and dynamic environments.
RoboTransfer~\cite{robotransfer} employs video diffusion models
conditioned on depth, surface normals, and object-level attributes to
re-generate demonstrations with modified visual textures, explicitly
targeting the sim-to-real and real-to-real texture gap in order to
improve policy performance.

Unlike these approaches, our Robotic Scene Cloning goes beyond texture
transfer: starting from real demonstration trajectories, we inject
deployment-specific objects, including moderate cross-shape
replacements, while preserving task-relevant geometry, thereby improving
the reuse of each trajectory across multiple target objects.

\begin{figure*}[h]
    \centering
       \includegraphics[width=1.0\textwidth]{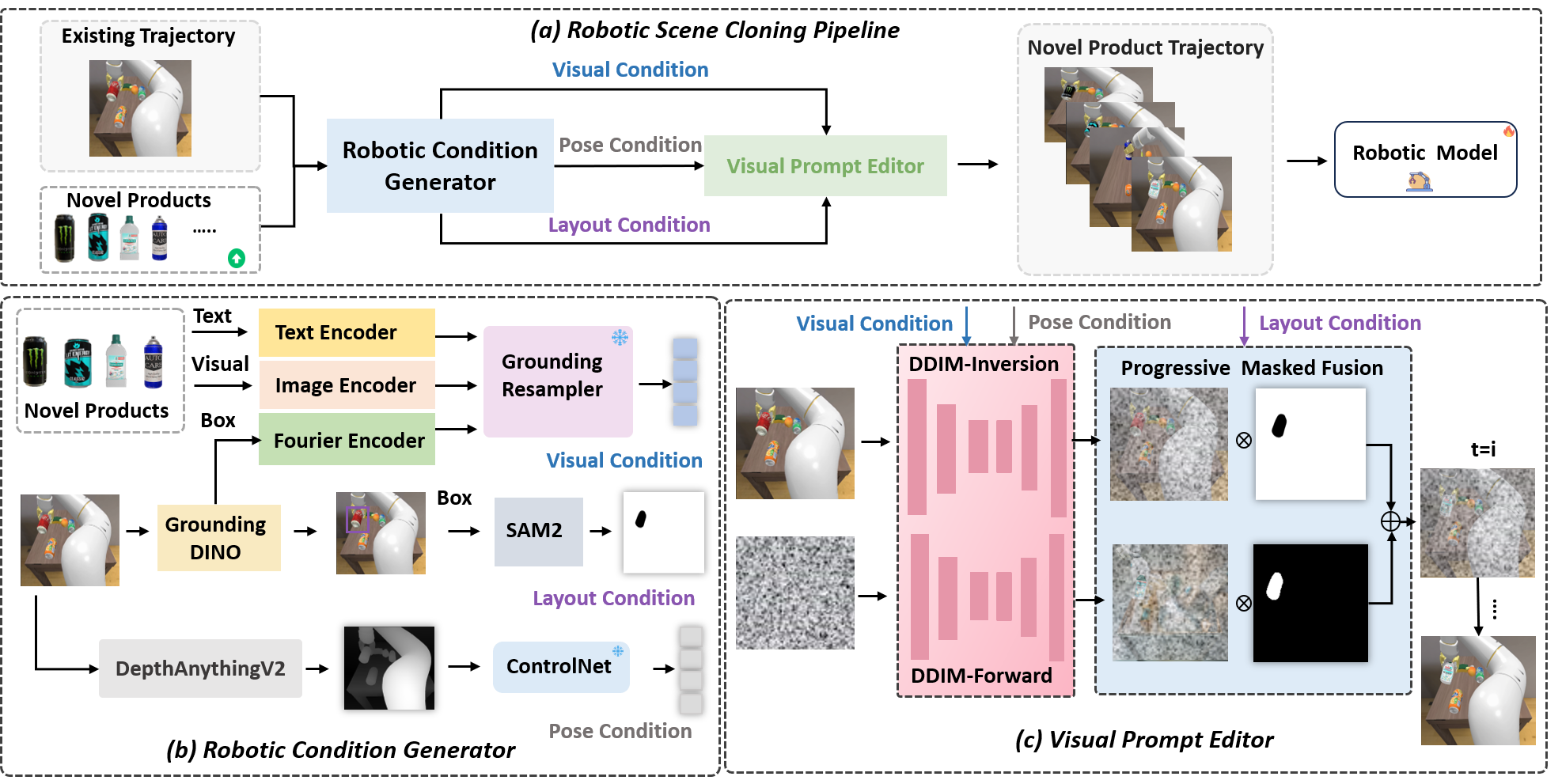}
    \caption{
Overview for \textbf{Robotic Scene Cloning}.
 \textbf{(a) RSC pipeline}  follows a two-stage process. Robotic Condition Generator prepares scene-specific conditions from training trajectories and a new product. Visual Prompt Editor then generates visually cloned trajectories, which fine-tune robotic models for better adaptation to novel products.
\textbf{(b) Robotic Condition Generator}prepares three conditions: First, the Grounding Resampler combines the new product's visual, textual, and positional encodings to generate position-bound visual conditions. Second, Grounding-DINO + SAM2 extract the coordinates and masks of existing products. Third, DepthAnythingV2 and ControlNet capture the pose conditions.
\textbf{(c) Visual Prompt Editor} applies the three conditions—visual, pose, and layout—generated by the Robotic Condition Generator. Visual and pose conditions guide the DDIM processes, while the layout condition is used in the progressive masked fusion.
}
    \label{fig:method}
\end{figure*}

\section{Method}
In this section, we introduce Robotic Scene Cloning (RSC), a novel data augmentation method designed for robotic scenarios, using visual prompts to address the zero-shot generalization challenge when deploying robotic policies to new environments.

\subsection{Pipeline of Robotic Scene Cloning}
We provide an overview of the RSC pipeline, which leverages MS-Diffusion~\cite{ms-diffusion}, a personalized concept generation model pretrained on large-scale personalization data, enabling visual generation with a single prompt image as visual prompt without per-object fine-tuning. While MS-Diffusion can generate novel visual concepts at specified locations, it has two key limitations for robotic data augmentation. First, it introduces deployment-irrelevant noise outside the target region, disrupting the semantic consistency required for robotic task deployment. To address this, we introduce the Visual Prompt Editor(Fig. \ref{fig:method} (c)), which injects new visual concepts while preserving the relevant context of the deployment environment. 
Second, 
MS-Diffusion only enforces coarse box-level layout and lacks explicit
pose constraints, so it does not control the grasp orientation of the
edited object. As a result, even if the object appears in the correct
region, its orientation may still be misaligned with a feasible or
reliable robot grasp.
To solve this, we propose the Robotic Condition Generator (Fig. \ref{fig:method} (b)), which precisely adjusts control conditions to ensure pose-aligned augmentations, allowing the robotic policy to adapt effectively to novel objects.

\subsubsection{Robotic Condition Generator}
\label{sec:rcg}

As shown in Fig. \ref{fig:method}(b), Robotic Condition Generator is designed to carefully prepare control conditions tailored for robotic scenarios, including visual, layout, and pose conditions for precise and 
high-quality data augmentation.

\textbf{Visual Condition.}  
We extract a visual-prompt embedding $f_v$ using the CLIP image encoder on a \emph{single} RGB image of the novel product (no background removal is required, and the prompt can be an in-the-wild product photo), a text-prompt embedding $f_t$ using the CLIP text encoder on the user description, and a grounding embedding $f_g$ by Fourier-encoding the bounding-box coordinates extracted by Grounding-DINO~\cite{grounding}. A learnable query embedding $f_q$ is initialized from $(f_t, f_g)$. These embeddings are then processed by the Grounding Resampler~\cite{ms-diffusion}, which uses an attention mechanism to combine them into the final visual condition:
\begin{equation}
c_{\text{visual}} = \mathrm{GroundingResampler}\bigl(f_v,\,f_q).
\end{equation}
The attention mechanism within the groundingresampler is described by:
\begin{equation}
\text{RSAttn} 
= \text{Softmax}\left( 
    \frac{\mathbf{Q}(f_q)\,\big(\mathbf{K}([f_v, f_q])\big)^{\mathsf T}}{\sqrt{d}} 
  \right)\,
  \mathbf{V}([f_v, f_q]).
\end{equation}
where \(\mathbf{Q}(f_q)\) is the query vector derived from \(f_q\), \(\mathbf{K}([f_v, f_q])\) is the concatenated key vector formed from the visual and query embeddings, and \(\mathbf{V}([f_v, f_q])\) is the corresponding value vector. 
This attention layer allows the model to enhance the interaction between the visual features and grounding features, injecting the position information to the visual condition.

\textbf{Layout Condition.}  
The layout condition \(c_{\text{layout}}\) is obtained by feeding the Grounding‐DINO bounding box into SAM2~\cite{grounding}, where the black regions in the resulting mask denote editable areas, while the white regions indicate parts of the image to be preserved unedited.

\textbf{Pose Condition.}  
To ensure pose-consistent augmentations—particularly correct object orientation—we pass the depth map from DepthAnythingV2~\cite{depthanythingv2} through ControlNet ~\cite{controlnet}:
\begin{equation}
c_{\text{pose}} = \mathrm{ControlNet}\bigl(\mathrm{DepthAnythingV2}(I)\bigr).
\end{equation}
The resulting depth‐conditioned signal inherently provided pose consistency when the Visual Prompt Editor generates new object in the specified location.

Robotic Condition Generator outputs the three independent control signals \((c_{\text{visual}},\,c_{\text{layout}},\,c_{\text{pose}})\), which are then fed—without further training—into the Visual Prompt Editor to drive visual-aligned, spatially precise, and pose-consistent data augmentation.

\subsection{Visual Prompt Editor}
As shown in Fig. \ref{fig:method}(c), Visual Prompt Editor injects a new visual concept into an existing scene while preserving the integrity of non-edited regions, enabling adaptation to novel deployment scenarios.

VPE integrates two complementary mechanisms:  
(1) \textbf{Progressive Masked Fusion}, which gradually merges stored inversion features with newly generated latents using a timestep-dependent blending mask, ensuring a smooth transition between edited and untouched regions; and  
(2) \textbf{Visual-Prompt Guided Image Editing}, where visual and pose conditions are injected to align generation with the desired spatial and pose cues.
By combining these mechanisms, Visual-Prompt Guided Image Editing avoids introducing deployment-irrelevant artifacts, guarantees spatially precise concept placement, and maintains high-fidelity reconstruction of unedited content. The following subsections describe Progressive Masked Fusion and Visual-Prompt-Guided Image Editing in detail.

\subsubsection{Progressive Masked Fusion}
Progressive Masked Fusion integrates a visual prompt into a target scene while preserving non-edited regions.  It achieves this by coupling the editing process with stored inverted latents from the original image in the \textbf{Inversion} stage, and progressively blending newly generated latents into targeted areas under spatial control in the \textbf{Denoising} stage.

\paragraph{Inversion}
The input image \(I\) is first encoded into the latent space \(\mathcal{Z}\) using a pretrained variational autoencoder (VAE)\cite{sdxl}, yielding an initial latent representation \(\mathbf{z}_0\).This latent representation serves as the starting point for recovering the image's diffusion trajectory. We then perform DDIM inversion over \(T\) timesteps to obtain the corresponding representation:
\begin{equation}
\mathbf{z}_t = \mathrm{DDIMInv}(\mathbf{z}_0, t),
\text{for }  t = 1, \dots, T
\end{equation}
We store each $\mathbf{z}_t$ as an \emph{anchor}, which preserves the content
of non-edited regions and enables faithful reconstruction during denoising,
while maintaining the visual fidelity of the original image as new elements
are added.

\paragraph{Denoising}
Starting from Gaussian noise $\tilde{\mathbf{z}}_T \sim \mathcal{N}(0, I)$, we iteratively denoise while injecting visual and pose conditions to generate new content in the target region. At each timestep $t$, we fuse the stored anchor $\mathbf{z}_t$ (representing the original image state) with the current denoised latent $\tilde{\mathbf{z}}_t$ (containing newly generated content) using a progressive blending strategy:
\begin{equation}
\begin{split}
\tilde{\mathbf{z}}_{t-1}
= \mathrm{DDIMForward}\!\left(
    M_t \odot \mathbf{z}_t + (1 - M_t) \odot \tilde{\mathbf{z}}_t,\; t
  \right),\\
\text{for } t = T, \dots, 1.
\end{split}
\end{equation}

where \(M_t\) is a linear-decay blending mask defined as:
\begin{equation}
M_t = c_{\text{layout}} \cdot \left(1 - \alpha \cdot \frac{T-t}{T}\right), 
\quad c_{\text{layout}}, \alpha \in [0,1]
\end{equation}
Here, $c_{\text{layout}}$ is the layout mask with values of 1 for regions to preserve and 0 for editable regions, and $\alpha$ controls the temporal decay rate of the blending process.
By gradually relaxing $M_t$ during denoising, the blending process preserves the overall scene context and spatial consistency at early stages, while allowing increasing flexibility for shape and appearance changes within the editable region at later stages.
This behavior is compatible with cross-shape editing scenarios, where the target object undergoes noticeable geometric changes while the surrounding scene remains unchanged.
The linear decay of $M_t$ further ensures a smooth transition at edit boundaries, enabling seamless integration of new content without introducing artifacts in non-edited areas.
Upon completion (\(t=0\)), the final latent \(\tilde{\mathbf{z}}_0\) is decoded by VAE to yield the edited image \(\hat{I} = \mathrm{Decode}(\tilde{\mathbf{z}}_0)\).

\subsubsection{Visual-Prompt Guided Image Editing}  
At each denoising step \(t\), we inject visual condition \(\mathbf{c}_{visual}\) via masked cross-attention and pose prior \(\mathbf{c}_{pose} \)  via ControlNet-guided modulation. 
These pose-aware features are then added as residuals to the intermediate
activations (e.g., the middle block and decoder blocks), which we denote
abstractly as:
\begin{equation}
\mathbf{z}_{\text{ctrl}} = \tilde{\mathbf{z}}_t + c_{\text{pose}}.
\end{equation}
where \(\mathbf{z}_{\text{ctrl}}\) denote the denoised latent with injected pose prior.
In the following process,  following\cite{ms-diffusion}, the masked cross-attention layers are applied to inject the visual prompt \(f_{\text{v}}\) to the denoised latents:
\begin{equation}
\mathbf{z}_{\text{visual}}
= \mathrm{Softmax}\left( \frac{Q K^\top}{\sqrt{d}} + M' \right) V.
\end{equation}
\begin{equation}
Q = W_q \mathbf{z}_{\text{ctrl}}^T, \quad
K = W_k c_{\text{visual}}^T, \quad
V = W_v c_{\text{visual}}^T
\end{equation}
where the mask \( M' \) restricts attention to editable regions defined by \(\mathbf{c}_{layout}\), assigning 0 to the editable areas  and \(-\infty\) to the non-editable areas. 
Combined with Progressive Masked Fusion, which blends condition-aware latents into editable areas while preserving anchor features elsewhere, this mechanism enables spatially precise and high-fidelity editing.

\subsection{Implementation Details}
The performance validation is divided into two parts: simulation and real-world testing. 

\textbf{Simulation Setting.}  
For simulation validation, we utilize SIMPLER\cite{simpler}, a real-world-aligned robotic evaluation environment and CALVIN\cite{calvin}, which focuses on long-horizon, language-conditioned robotic task.
In SIMPLER, we adopt CogACT\cite{cogact}, a state-of-the-art policy learning baseline, as our primary method.
To assess the impact of different augmentation strategies, we first employ CogACT to collect 67 successful grasping trajectories of a Coke can. We then apply two augmentation techniques to edit the original Coke can trajectories: GreenAug, which uses text prompts to augment robotic data, and our proposed RSC, which employs visual prompting for scene cloning.
We fine-tune CogACT on these augmented datasets using a global batch size of 256, 8 diffusion steps per sample, and a learning rate of 2e-5. Training is performed on 8 NVIDIA A100 GPUs, with evaluations conducted on a single NVIDIA A100 GPU. We varied distractor types or object positions during testing to assess if the policy captures visual features consistent with the instructions. Each tasks is tested 10 times horizontally and 10 times vertically, for 20 total trials.
In CALVIN, we utilize four distinct manipulation environments. Each environment involves a robotic arm interacting with objects such as blocks, drawers and sliders. The benchmark supports 34 unique tasks, challenging agents to follow sequential language instructions to perform complex manipulation actions. For policy learning, we adopt RoboFlamingo \cite{robofg}, an advanced method from the CALVIN benchmark.

\begin{figure}[t]
    \centering
              \includegraphics[width=0.5\textwidth]{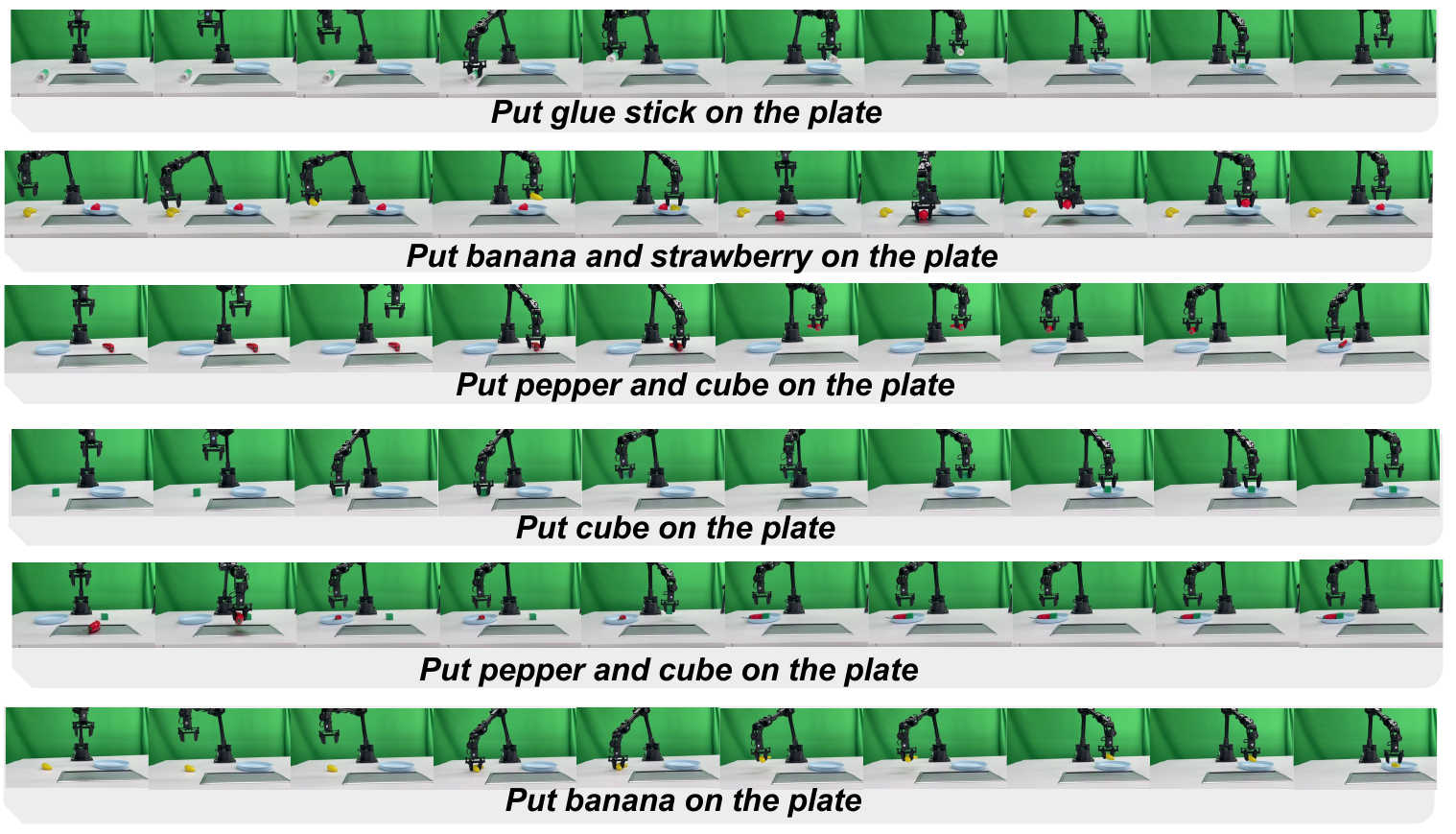}
    \caption{
The visualization results of our real-world validation task on the WidowX robot.
}
    \label{fig:real-task}
\end{figure}

\textbf{Real-World Setting.} For real-world experiments, we utilize the WidowX250S robot, equipped with an Intel RealSense D435 camera mounted in front to provide third-person visual input. Our baseline model\cite{rosa}, a Vision-Language-Action (VLA) model based on the LLaVA \cite{llava} architecture, integrates the Qwen-2.5-7B model \cite{qwen2} as the large language model backbone, CLIP ViT-L/14 \cite{clip} as the vision encoder, and a two-layer MLP as the projector. The model was pre-trained on the Cambrian 737k dataset \cite{cambrian}, combined with real-world physical robotic data, on 8 A100 GPUs, using a learning rate of 2e-5, a warmup cosine schedule \cite{warmup}, a batch size of 64, and a tokenizer range of 256, and shows comparable performance to CogACT in our real-world testing. 
The model is evaluated on two \emph{Real-Demo Task} and four \emph{No-Real-Demo Task} to assess generalization.
Tasks are categorized strictly by the availability of \emph{real-world robot demonstrations}: real-demo tasks provide demonstrations used for fine-tuning, while no-real-demo tasks have no real demonstrations; RSC-edited trajectories are treated as synthetic augmentations and do not alter this categorization.
For fine-tuning, we consider two training sets: Baseline uses only the original real-world demonstrations from the real-demo tasks, whereas Baseline+RSC uses the same demonstrations together with additional RSC-edited trajectories derived from them.
We fine-tune one multi-task model for Baseline on the combined real-demo demonstrations, and one for Baseline+RSC on the same real-demo demonstrations augmented with RSC-edited trajectories, and evaluate each model across all tasks (no per-task training).
The real-demo tasks include a single-object task (placing a banana on a plate) and a multi-object task (placing a banana and a strawberry on a plate).
The no-real-demo tasks involve placing a cube on a plate, placing a pepper on a plate, placing a glue stick on a plate, and placing both a pepper and a cube on a plate. 
Examples of these tasks are illustrated in Fig. \ref{fig:real-task}. Each task was evaluated over 10 trials, consistent with prior experimental protocols \cite{openvla,pai0}.

\begin{figure}[t]
    \centering
          \includegraphics[width=0.5\textwidth]{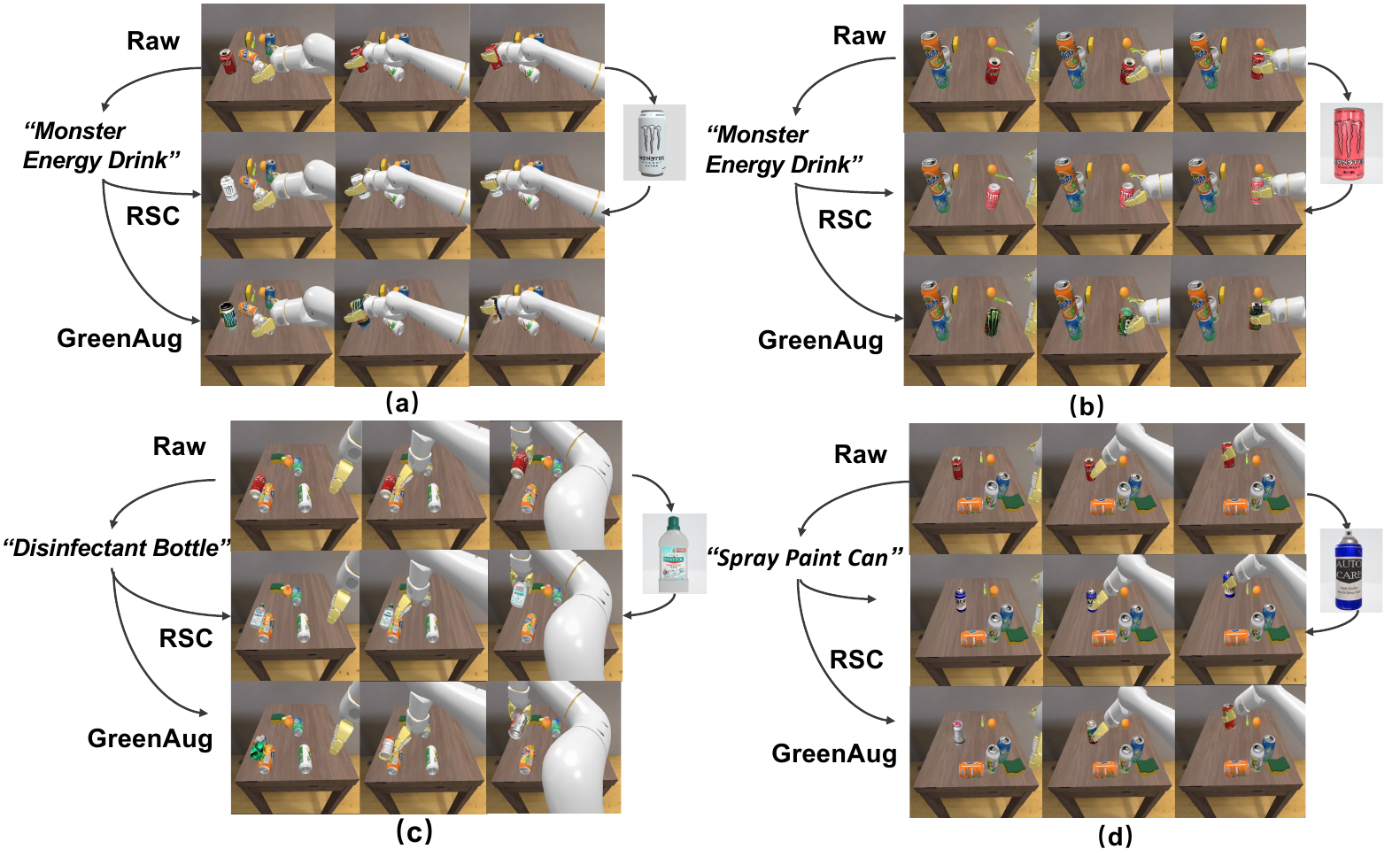}xx
    \caption{
Editing effect comparison between RSC and GreenAug in the SIMPLER Benchmark.
}
    \label{fig:simpler-vis}
\end{figure}

\begin{figure*}[htbp]
    \centering
    \includegraphics[width=0.85\textwidth]{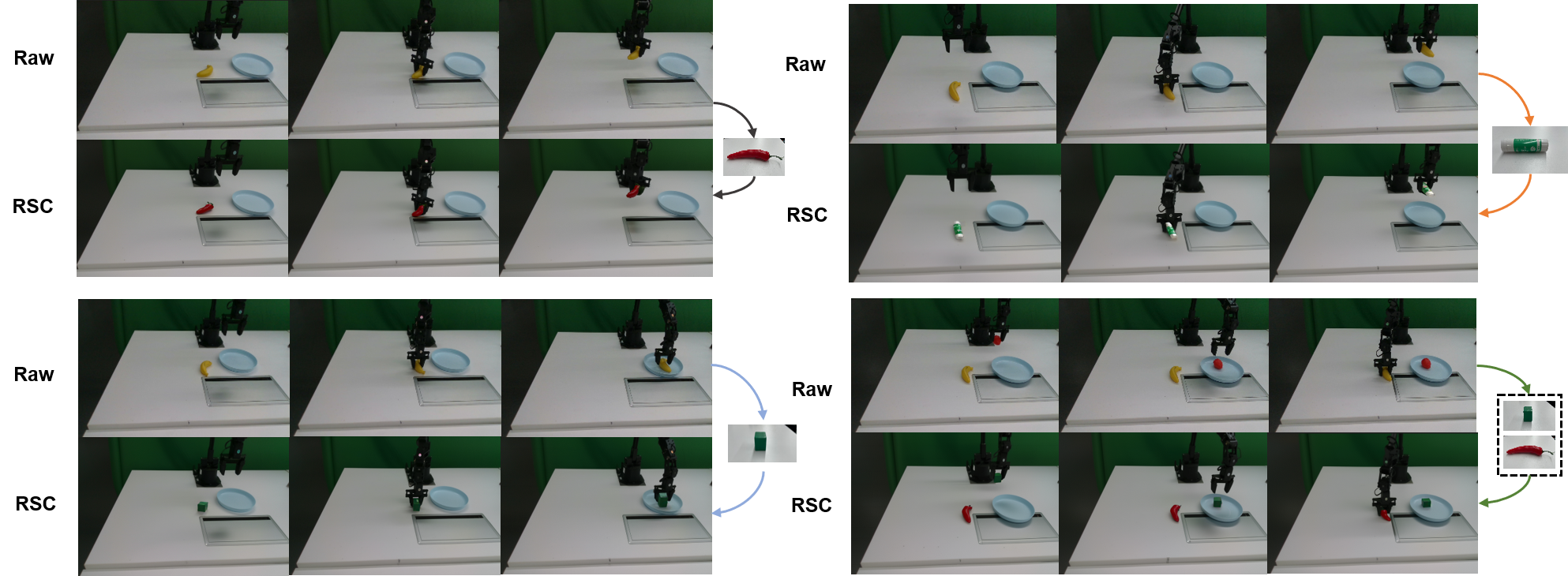}
    \caption{
Each pair of rows shows the original video frames (Raw) and the corresponding frames edited by our method (RSC) after replacing the target object. 
}
    \label{fig:real-vis}
\end{figure*}

\begin{table*}[t]
\centering
\small
\begin{tabular}{lccccc}
\toprule
Method 
  & MonsterEnergy Drink V1
  & MonsterEnergy Drink V2
  & Disinfectant Bottle
  & Spray Paint Can
  & \emph{Avg} \\
\midrule
OpenVLA\cite{openvla} 
  & 0.0  & 0.0  & 0.0  & 0.0  & 0.0  \\ 
CogACT\cite{cogact}   
  & 0.0  & 0.0  & 45.0 & 10.0 & 13.8 \\
CogACT\cite{cogact}+GAug\cite{greenaug}
  & 5.0  & 10.0 & 45.0 & 25.0 & 21.3 \\
CogACT\cite{cogact}+RSC                  
  & \textbf{70.0} & \textbf{50.0} & \textbf{60.0} 
  & \textbf{45.0} & \textbf{56.3} \\
\bottomrule
\end{tabular}
\vspace{0.1cm}
\caption{Performance of state-of-the-art VLA models on four novel deployment tasks of the SIMPLER benchmark, highlighting the impact of GreenAug (GAug) and Robotic Scene Cloning.}
\label{tab:simpler}
\end{table*}

\begin{table*}[t]
\centering
\small
\begin{tabular}{lcccccc}
\toprule
\multirow{2}{*}{Method} & \multicolumn{4}{c}{No-Real-Demo Task} & \multicolumn{2}{c}{Real-Demo Task} \\
\cmidrule(lr){2-5} \cmidrule(lr){6-7}
 & Multi-Object & \multicolumn{3}{c}{Single Object} & Multi-Object & Single Object \\
\midrule
 & Pepper \& Cube & Cube & Glue stick & Pepper & Banana \& Strawberry & Banana \\
\midrule
Baseline\cite{rosa} & 20\% & 40\% & 40\% & 40\% & 50\% & 70\% \\
Baseline\cite{rosa} +RSC &\textbf{50\%} & \textbf{80\%}  &\textbf{60\%}  &\textbf{70\%}  & 50\% & 70\% \\
\bottomrule
\end{tabular}
  \vspace{0.1cm}
\caption{
We evaluate two \emph{real-demo tasks} and four \emph{no-real-demo tasks}.
Real-demo tasks include a single-object task (banana) and a multi-object task (banana \& strawberry), both of which provide real-world robot demonstrations used for fine-tuning.
No-real-demo tasks involve novel objects or object combinations (cube, glue stick, pepper, and pepper \& cube) and provide no real-world robot demonstrations; they are used solely for evaluation.
All results report success rates over 10 trials per task.
}
\label{tab:real_tab}
\end{table*}

\subsection{Experiment  Results on Realistic Robotic Benchmark (SIMPLER).}
In this section, we explore two meaningful application scenarios for robotic scene cloning in SIMPLER benchmark\cite{simpler}: cross-texture and cross-shape product cloning.

\textbf{Cross-Texture Product Cloning.}
Cross-texture cloning helps robots quickly adapt to new products in real-world deployment. For example, In automated supermarket sorting tasks, existing policies are traditionally limited to familiar products and struggle with new ones. In this case, by injecting visual concept of novel product by RSC into existing trajectories can improve generalization without requiring new data collection. As shown in Table \ref{tab:simpler}, OpenVLA and CogACT exhibit limited zero-shot generalization to new beverage types (e.g., two Monster Energy variants), far from meeting real-world application needs.
While existing augmentation methods such as GreenAug improve performance to around 10\%, their effectiveness remains limited. In contrast, RSC achieves an average performance of approximately 60\% using visual prompts with the same amount of data, substantially enhancing the generalization of robotic policies to new scenarios. Visual comparisons between GreenAug and RSC in Fig. \ref{fig:simpler-vis}(a)(b) further highlight the precise visual instruction-following capability of RSC.

\textbf{Cross-Shape Product Cloning.}
In real-world scenarios, product cloning needs extend beyond visual textures. There are cases where the shapes of objects in the existing training dataset do not strictly align with those of new products.
For example, in industrial automation, robots may need to handle disinfectant bottles for equipment sanitization or spray paint cans for metal coating. However, if a robotic policy has only been trained on interactions with Coke bottles, it lacks familiarity with these new objects. In such cases, cross-shape visual cloning becomes essential.
As shown in Table \ref{tab:simpler}, our method enables effective cross-shape visual cloning by reducing ControlNet’s control scale. Furthermore, Table \ref{tab:simpler} shows that our approach improves the SOTA policy by approximately 30\% on these objects, substantially outperforming the text-prompted embodied augmentation method GreenAug, which achieves only 7.5\%. Visual comparisons in Fig. \ref{fig:simpler-vis}(c)(d) further illustrate the cross-shape editing capability of our method, going beyond what text-prompt-based approaches can achieve.

\subsection{Experiment Results on Real-World Scenarios.}
As shown in Table \ref{tab:real_tab}, our real-world experiments report the performance of the Baseline and Baseline+RSC models on both No-Real-Demo and Real-Demo tasks. To highlight the broad adaptability of RSC, we evaluate tasks ranging from short-range single-object manipulation to long-range multi-object manipulation.
Additionally, the shapes of the objects in each task category were varied to further assess RSC's effectiveness in cross-shape cloning.

\textbf{Single-Object, Short-Range, Cross-Shape Task Cloning:}
In this setup, RSC utilizes the trajectory of a single-object, short-range task (putting a banana on the plate) as the primary data source.
A single image of a new object (cube, glue stick, or pepper) is provided as a visual prompt and injected at the banana’s position in the original trajectory to generate grasping trajectories for the new objects.
Visualization results are shown in Fig.~\ref{fig:real-vis}.
As reported in Table~\ref{tab:real_tab}, augmenting the training set with RSC-enhanced grasping trajectories improves the Baseline’s performance by 40\% on the ``put the cube on the plate'' task, 30\% on the ``put the pepper on the plate'' task, and 20\% on the ``put the glue stick on the plate'' task.
These results indicate that RSC improves the model’s ability to generalize to tasks without real-world demonstrations, including object-level shape changes relative to the original banana (e.g., adapting a banana-putting trajectory to a cube or glue stick).
Importantly, RSC maintains stable performance on the seen single-object task (banana).

\textbf{Multi-Object, Long-Range, Cross-Shape Task Cloning}: 
In this configuration, RSC leverages the trajectory of a long-range, multi-object task (sequentially placing a banana and a strawberry on a plate) as the primary data source. A single image containing a pepper and a cube is used as the visual prompt, which is injected at the positions of the original objects in the trajectory to generate manipulation trajectories for the new objects.
Visualization results are presented in Fig.~\ref{fig:real-vis}.
As shown in Table~\ref{tab:real_tab}, incorporating RSC-enhanced manipulation trajectories improves the Baseline’s performance by 30\% on the long-range, multi-object task of sequentially placing a cube and a pepper on the plate, a task without real-world demonstrations.
Additionally, RSC maintains stable performance on the real-demo multi-object task (banana and strawberry).

\subsection{Experiment Results in Long-Horizon Robotic Benchmark(CALVIN).}

\begin{figure}[h]
\centering
    \includegraphics[width=0.4\textwidth]{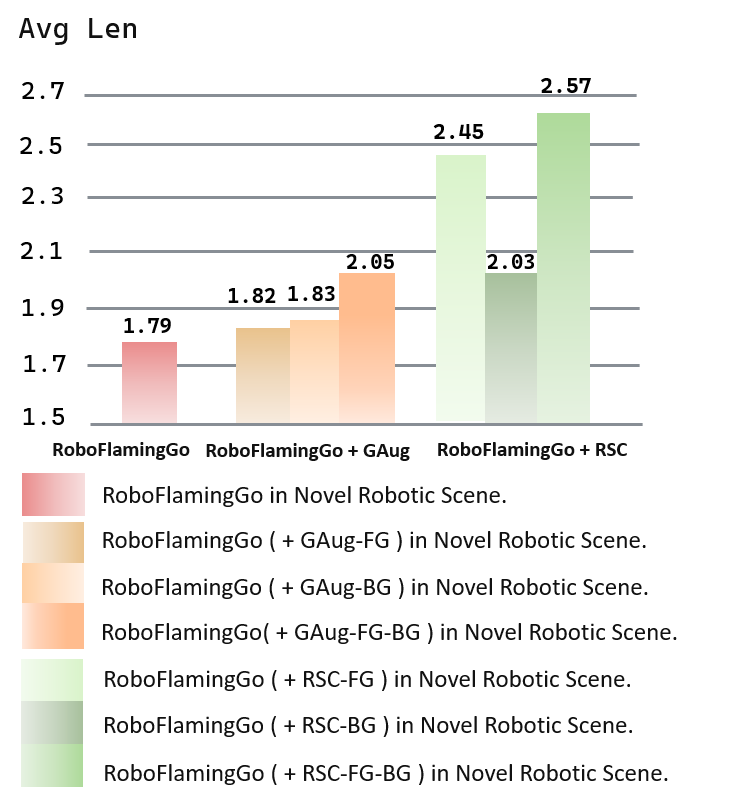}
    \caption{
Comparison of different data augmentation strategies on the CALVIN long-horizon task. AvgLen represents the average task execution length. FG denotes foreground augmentation, and BG denotes background augmentation.
}
    \label{fig:calvin}
\end{figure} 

We further analyze how foreground and background augmentation jointly
contribute to scene-level transfer on long-horizon tasks in CALVIN benchmark.
 Here, the
\emph{foreground} denotes the manipulable objects whose appearance can
change (e.g., the color or texture of the block to be manipulated), while
the \emph{background} denotes the surrounding scene appearance (e.g.,
the table color and wood texture).

\textbf{Joint Foreground–Background Augmentation for Scene Cloning.}
Using RoboFlamingo as the base policy, Fig. \ref{fig:calvin} compares foreground-only
(FG), background-only (BG), and joint FG+BG augmentation implemented
with either GreenAug or RSC. The baseline reaches an average successful
sequence length of only 1.79 in novel scenes where both the objects and
the table appearance are unseen. GreenAug improves this to 1.82/1.83
with FG/BG alone and to 2.05 with joint FG+BG. In contrast, RSC, which
clones both the object and table appearance via visual prompts,
achieves 2.03/2.45 with FG/BG alone and 2.57 when cloning the full
scene (FG+BG). This decomposition shows that explicitly modeling a
scene as “foreground + background” and cloning both components jointly
provides the strongest gains in scene transferability.

\section{Conclusion and Limitation.}
In this paper, we introduce RSC, a new data synthesis method to boost zero-shot deployment of robotic models by improving adaptation to new environments. Unlike traditional augmentation methods focused on general diversity, RSC uses visual cues to create scene-consistent synthetic data, enabling precise adaptation to deployment settings while cutting data collection costs.
Experiments show that in simulation environments like SIMPLER, RSC outperforms leading embodied data augmentation methods in cross-texture and cross-shape generalization tasks for new objects. In real-world tests, RSC also excels in challenging multi-object and cross-shape tasks, proving its practicality and versatility.
These findings highlight RSC’s potential to bridge the training-deployment gap in embodied intelligence tasks. However, RSC is currently limited to moderate shape variations, struggling with significant shape changes. Fine-tuning with large-scale personalized datasets could improve its adaptability to diverse object shapes.
This work offers a starting point for tackling robotic generalization in specific deployment scenarios, providing a scalable alternative to expensive real-world data collection. We hope it encourages further research into data-efficient robotic learning and deployment.

\bibliographystyle{IEEEtran}
\bibliography{main}

@article{qwen2,
  title={Qwen2. 5 technical report},
  author={Yang, An and Yang, Baosong and Zhang, Beichen and Hui, Binyuan and Zheng, Bo and Yu, Bowen and Li, Chengyuan and Liu, Dayiheng and Huang, Fei and Wei, Haoran and others},
  journal={arXiv preprint arXiv:2412.15115},
  year={2024}
}

@article{simpler,
  title={Evaluating real-world robot manipulation policies in simulation},
  author={Li, Xuanlin and Hsu, Kyle and Gu, Jiayuan and Pertsch, Karl and Mees, Oier and Walke, Homer Rich and Fu, Chuyuan and Lunawat, Ishikaa and Sieh, Isabel and Kirmani, Sean and others},
  journal={arXiv preprint arXiv:2405.05941},
  year={2024}
}

@article{cambrian,
  title={Cambrian-1: A fully open, vision-centric exploration of multimodal llms},
  author={Tong, Peter and Brown, Ellis and Wu, Penghao and Woo, Sanghyun and IYER, Adithya Jairam Vedagiri and Akula, Sai Charitha and Yang, Shusheng and Yang, Jihan and Middepogu, Manoj and Wang, Ziteng and others},
  journal={Advances in Neural Information Processing Systems},
  volume={37},
  pages={87310--87356},
  year={2024}
}

@article{warmup,
  title={Sgdr: Stochastic gradient descent with warm restarts},
  author={Loshchilov, Ilya and Hutter, Frank},
  journal={arXiv preprint arXiv:1608.03983},
  year={2016}
}

@article{rt1,
  title={Rt-1: Robotics transformer for real-world control at scale},
  author={Brohan, Anthony and Brown, Noah and Carbajal, Justice and Chebotar, Yevgen and Dabis, Joseph and Finn, Chelsea and Gopalakrishnan, Keerthana and Hausman, Karol and Herzog, Alex and Hsu, Jasmine and others},
  journal={arXiv preprint arXiv:2212.06817},
  year={2022}
}

@article{roise,
  title={Scaling robot learning with semantically imagined experience},
  author={Yu, Tianhe and Xiao, Ted and Stone, Austin and Tompson, Jonathan and Brohan, Anthony and Wang, Su and Singh, Jaspiar and Tan, Clayton and Peralta, Jodilyn and Ichter, Brian and others},
  journal={arXiv preprint arXiv:2302.11550},
  year={2023}
}

@article{genaug,
  title={Genaug: Retargeting behaviors to unseen situations via generative augmentation},
  author={Chen, Zoey and Kiami, Sho and Gupta, Abhishek and Kumar, Vikash},
  journal={arXiv preprint arXiv:2302.06671},
  year={2023}
}

@article{greenaug,
  title={Green Screen Augmentation Enables Scene Generalisation in Robotic Manipulation},
  author={Teoh, Eugene and Patidar, Sumit and Ma, Xiao and James, Stephen},
  journal={arXiv preprint arXiv:2407.07868},
  year={2024}
}

@article{cogact,
  title={CogACT: A Foundational Vision-Language-Action Model for Synergizing Cognition and Action in Robotic Manipulation},
  author={Li, Qixiu and Liang, Yaobo and Wang, Zeyu and Luo, Lin and Chen, Xi and Liao, Mozheng and Wei, Fangyun and Deng, Yu and Xu, Sicheng and Zhang, Yizhong and others},
  journal={arXiv preprint arXiv:2411.19650},
  year={2024}
}

@article{llava,
  title={Visual instruction tuning},
  author={Liu, Haotian and Li, Chunyuan and Wu, Qingyang and Lee, Yong Jae},
  journal={Advances in neural information processing systems},
  volume={36},
  year={2024}
}

@article{roboengine,
  title={RoboEngine: Plug-and-Play Robot Data Augmentation with Semantic Robot Segmentation and Background Generation},
  author={Yuan, Chengbo and Joshi, Suraj and Zhu, Shaoting and Su, Hang and Zhao, Hang and Gao, Yang},
  journal={arXiv preprint arXiv:2503.18738},
  year={2025}
}

@article{calvin,
  title={Calvin: A benchmark for language-conditioned policy learning for long-horizon robot manipulation tasks},
  author={Mees, Oier and Hermann, Lukas and Rosete-Beas, Erick and Burgard, Wolfram},
  journal={IEEE Robotics and Automation Letters},
  volume={7},
  number={3},
  pages={7327--7334},
  year={2022},
  publisher={IEEE}
}

@article{robotransfer,
  title={RoboTransfer: Geometry-Consistent Video Diffusion for Robotic Visual Policy Transfer},
  author={Liu, Liu and Wang, Xiaofeng and Zhao, Guosheng and Li, Keyu and Qin, Wenkang and Qiu, Jiaxiong and Zhu, Zheng and Huang, Guan and Su, Zhizhong},
  journal={arXiv preprint arXiv:2505.23171},
  year={2025}
}

@article{ms-diffusion,
  title={MS-Diffusion: Multi-subject Zero-shot Image Personalization with Layout Guidance},
  author={Wang, X and Fu, Siming and Huang, Qihan and He, Wanggui and Jiang, Hao},
  journal={arXiv preprint arXiv:2406.07209},
  year={2024}
}

@article{grounding,
  title={Grounding dino: Marrying dino with grounded pre-training for open-set object detection},
  author={Liu, Shilong and Zeng, Zhaoyang and Ren, Tianhe and Li, Feng and Zhang, Hao and Yang, Jie and Li, Chunyuan and Yang, Jianwei and Su, Hang and Zhu, Jun and others},
  journal={arXiv preprint arXiv:2303.05499},
  year={2023}
}

@article{depthanythingv2,
  title={Depth Anything V2},
  author={Yang, Lihe and Kang, Bingyi and Huang, Zilong and Zhao, Zhen and Xu, Xiaogang and Feng, Jiashi and Zhao, Hengshuang},
  journal={arXiv preprint arXiv:2406.09414},
  year={2024}
}

@article{rt2,
  title={Rt-2: Vision-language-action models transfer web knowledge to robotic control},
  author={Brohan, Anthony and Brown, Noah and Carbajal, Justice and Chebotar, Yevgen and Chen, Xi and Choromanski, Krzysztof and Ding, Tianli and Driess, Danny and Dubey, Avinava and Finn, Chelsea and others},
  journal={arXiv preprint arXiv:2307.15818},
  year={2023}
}

@inproceedings{peract,
  title={Perceiver-actor: A multi-task transformer for robotic manipulation},
  author={Shridhar, Mohit and Manuelli, Lucas and Fox, Dieter},
  booktitle={Conference on Robot Learning},
  pages={785--799},
  year={2023},
  organization={PMLR}
}

@article{robofg,
  title={Vision-language foundation models as effective robot imitators},
  author={Li, Xinghang and Liu, Minghuan and Zhang, Hanbo and Yu, Cunjun and Xu, Jie and Wu, Hongtao and Cheang, Chilam and Jing, Ya and Zhang, Weinan and Liu, Huaping and others},
  journal={arXiv preprint arXiv:2311.01378},
  year={2023}
}

@inproceedings{cliport,
  title={Cliport: What and where pathways for robotic manipulation},
  author={Shridhar, Mohit and Manuelli, Lucas and Fox, Dieter},
  booktitle={Conference on robot learning},
  pages={894--906},
  year={2022},
  organization={PMLR}
}

@article{mandi2022cacti,
  title={Cacti: A framework for scalable multi-task multi-scene visual imitation learning},
  author={Mandi, Zhao and Bharadhwaj, Homanga and Moens, Vincent and Song, Shuran and Rajeswaran, Aravind and Kumar, Vikash},
  journal={arXiv preprint arXiv:2212.05711},
  year={2022}
}

@article{sdxl,
  title={Sdxl: Improving latent diffusion models for high-resolution image synthesis},
  author={Podell, Dustin and English, Zion and Lacey, Kyle and Blattmann, Andreas and Dockhorn, Tim and M{\"u}ller, Jonas and Penna, Joe and Rombach, Robin},
  journal={arXiv preprint arXiv:2307.01952},
  year={2023}
}

@article{rosa,
  title={ROSA: Harnessing Robot States for Vision-Language and Action Alignment},
  author={Wen, Yuqing and Gu, Kefan and Liu, Haoxuan and Zhao, Yucheng and Wang, Tiancai and Fan, Haoqiang and Sun, Xiaoyan},
  journal={arXiv preprint arXiv:2506.13679},
  year={2025}
}

@inproceedings{clip,
  title={Learning transferable visual models from natural language supervision},
  author={Radford, Alec and Kim, Jong Wook and Hallacy, Chris and Ramesh, Aditya and Goh, Gabriel and Agarwal, Sandhini and Sastry, Girish and Askell, Amanda and Mishkin, Pamela and Clark, Jack and others},
  booktitle={International conference on machine learning},
  pages={8748--8763},
  year={2021},
  organization={PMLR}
}

@inproceedings{controlnet,
  title={Adding conditional control to text-to-image diffusion models},
  author={Zhang, Lvmin and Rao, Anyi and Agrawala, Maneesh},
  booktitle={Proceedings of the IEEE/CVF International Conference on Computer Vision},
  pages={3836--3847},
  year={2023}
}

@article{diffusionpolicy,
  title={Diffusion policy: Visuomotor policy learning via action diffusion},
  author={Chi, Cheng and Xu, Zhenjia and Feng, Siyuan and Cousineau, Eric and Du, Yilun and Burchfiel, Benjamin and Tedrake, Russ and Song, Shuran},
  journal={The International Journal of Robotics Research},
  pages={02783649241273668},
  year={2023},
  publisher={SAGE Publications Sage UK: London, England}
}

@article{r3m,
  title={R3m: A universal visual representation for robot manipulation},
  author={Nair, Suraj and Rajeswaran, Aravind and Kumar, Vikash and Finn, Chelsea and Gupta, Abhinav},
  journal={arXiv preprint arXiv:2203.12601},
  year={2022}
}

@article{openworld,
  title={Open-world object manipulation using pre-trained vision-language models},
  author={Stone, Austin and Xiao, Ted and Lu, Yao and Gopalakrishnan, Keerthana and Lee, Kuang-Huei and Vuong, Quan and Wohlhart, Paul and Kirmani, Sean and Zitkovich, Brianna and Xia, Fei and others},
  journal={arXiv preprint arXiv:2303.00905},
  year={2023}
}

@article{octo,
  title={Octo: An open-source generalist robot policy},
  author={Team, Octo Model and Ghosh, Dibya and Walke, Homer and Pertsch, Karl and Black, Kevin and Mees, Oier and Dasari, Sudeep and Hejna, Joey and Kreiman, Tobias and Xu, Charles and others},
  journal={arXiv preprint arXiv:2405.12213},
  year={2024}
}

@article{gr1,
  title={Unleashing large-scale video generative pre-training for visual robot manipulation},
  author={Wu, Hongtao and Jing, Ya and Cheang, Chilam and Chen, Guangzeng and Xu, Jiafeng and Li, Xinghang and Liu, Minghuan and Li, Hang and Kong, Tao},
  journal={arXiv preprint arXiv:2312.13139},
  year={2023}
}

@article{openvla,
  title={OpenVLA: An Open-Source Vision-Language-Action Model},
  author={Kim, Moo Jin and Pertsch, Karl and Karamcheti, Siddharth and Xiao, Ted and Balakrishna, Ashwin and Nair, Suraj and Rafailov, Rafael and Foster, Ethan and Lam, Grace and Sanketi, Pannag and others},
  journal={arXiv preprint arXiv:2406.09246},
  year={2024}
}

@article{pai0,
  title={$pi\_0 $: A Vision-Language-Action Flow Model for General Robot Control},
  author={Black, Kevin and Brown, Noah and Driess, Danny and Esmail, Adnan and Equi, Michael and Finn, Chelsea and Fusai, Niccolo and Groom, Lachy and Hausman, Karol and Ichter, Brian and others},
  journal={arXiv preprint arXiv:2410.24164},
  year={2024}
}

@article{aldm,
  title={ALDM-Grasping: Diffusion-aided Zero-Shot Sim-to-Real Transfer for Robot Grasping},
  author={Li, Yiwei and Wu, Zihao and Zhao, Huaqin and Yang, Tianze and Liu, Zhengliang and Shu, Peng and Sun, Jin and Parasuraman, Ramviyas and Liu, Tianming},
  journal={arXiv preprint arXiv:2403.11459},
  year={2024}
}

@article{cyclegan,
  title={Cyclegan, a master of steganography},
  author={Chu, Casey and Zhmoginov, Andrey and Sandler, Mark},
  journal={arXiv preprint arXiv:1712.02950},
  year={2017}
}

@inproceedings{lucidsim,
  title={Learning Visual Parkour from Generated Images},
  author={Yu, Alan and Yang, Ge and Choi, Ran and Ravan, Yajvan and Leonard, John and Isola, Phillip},
  booktitle={8th Annual Conference on Robot Learning},
  year={2024}
}

@inproceedings{rl-cyclegan,
  title={Rl-cyclegan: Reinforcement learning aware simulation-to-real},
  author={Rao, Kanishka and Harris, Chris and Irpan, Alex and Levine, Sergey and Ibarz, Julian and Khansari, Mohi},
  booktitle={Proceedings of the IEEE/CVF Conference on Computer Vision and Pattern Recognition},
  pages={11157--11166},
  year={2020}
}

@article{robo1,
  title={Palm-e: An embodied multimodal language model},
  author={Driess, Danny and Xia, Fei and Sajjadi, Mehdi SM and Lynch, Corey and Chowdhery, Aakanksha and Ichter, Brian and Wahid, Ayzaan and Tompson, Jonathan and Vuong, Quan and Yu, Tianhe and others},
  journal={arXiv preprint arXiv:2303.03378},
  year={2023}
}

@inproceedings{robo2,
  title={Lm-nav: Robotic navigation with large pre-trained models of language, vision, and action},
  author={Shah, Dhruv and Osi{\'n}ski, B{\l}a{\.z}ej and Levine, Sergey and others},
  booktitle={Conference on robot learning},
  pages={492--504},
  year={2023},
  organization={PMLR}
}

@article{robo3,
  title={Llama: Open and efficient foundation language models},
  author={Touvron, Hugo and Lavril, Thibaut and Izacard, Gautier and Martinet, Xavier and Lachaux, Marie-Anne and Lacroix, Timoth{\'e}e and Rozi{\`e}re, Baptiste and Goyal, Naman and Hambro, Eric and Azhar, Faisal and others},
  journal={arXiv preprint arXiv:2302.13971},
  year={2023}
}

@inproceedings{robo4,
  title={Robot learning with sensorimotor pre-training},
  author={Radosavovic, Ilija and Shi, Baifeng and Fu, Letian and Goldberg, Ken and Darrell, Trevor and Malik, Jitendra},
  booktitle={Conference on Robot Learning},
  pages={683--693},
  year={2023},
  organization={PMLR}
}
\end{document}